\documentclass[10pt,twocolumn,letterpaper]{article}

\usepackage[pagenumbers]{cvpr} 

\definecolor{cvprblue}{rgb}{0.21,0.49,0.74}
\usepackage[pagebackref,breaklinks,colorlinks,allcolors=cvprblue]{hyperref}
\usepackage{lipsum}
\usepackage[table]{xcolor}
\usepackage{enumitem}
\usepackage{algorithm}
\usepackage{algorithmicx}
\usepackage{algcompatible}
\usepackage{algpseudocode}
\usepackage{amsmath}
\usepackage{comment}

\def\paperID{11486} 
\def\confName{CVPR}
\def\confYear{2026}

\title{Thinking with Drafts: Speculative Temporal Reasoning for \\Efficient Long Video Understanding}

\author{
    \textbf{Pengfei Hu}$^{1,2*}$,
    \textbf{Meng Cao}$^{1,2*}$,
    \textbf{Yingyao Wang}$^{2*}$,
    \textbf{Yi Wang}$^{3}$,
    \textbf{Jiahua Dong}$^{1}$,\\
    \textbf{Jun Song}$^{2\dagger}$,
    \textbf{Yu Cheng}$^{2}$,
    \textbf{Bo Zheng}$^{2}$,
    \textbf{Xiaodan Liang}$^{1\dagger}$\\
    \\
    $^{1}$MBZUAI \quad
    $^{2}$Alibaba Group Holding Limited \quad
    $^{3}$Shanghai AI Lab \quad \\
    {\small $^*$Equal contribution \quad $^\dagger$Corresponding author}
}

\begin{document}

\twocolumn[{
	\renewcommand\twocolumn[1][]{#1}
	\maketitle
	\begin{center}
		\includegraphics[width=0.90\linewidth]{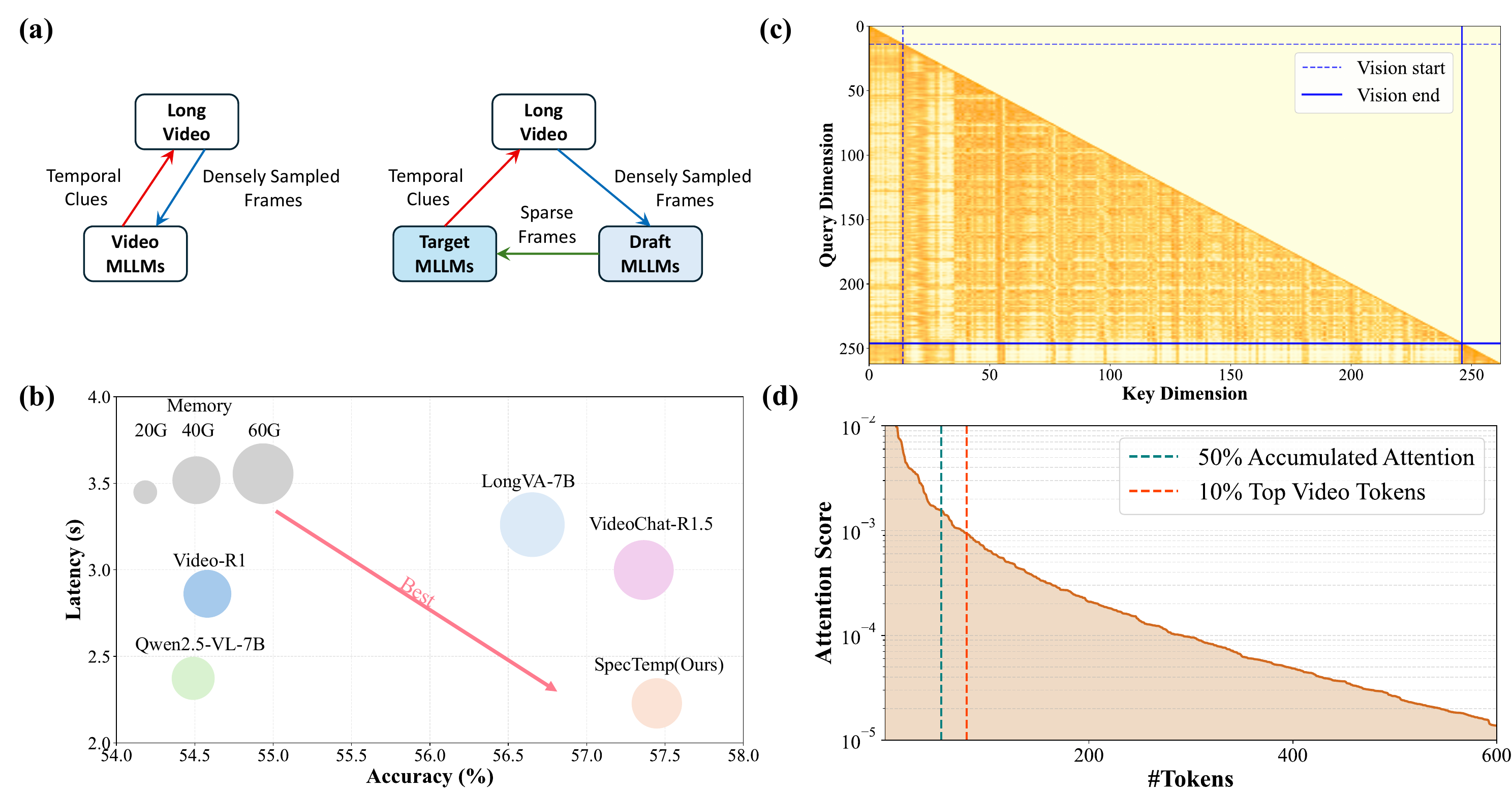}
		\captionsetup{type=figure}
     \caption{(a) \textbf{Paradigm comparison}. Left: Conventional MLLMs with growing multi-modal contexts. Right: SpecTemp's dual-model design—Target MLLM (7B) for reasoning, Draft MLLM (3B) for frame selection; (b) \textbf{Efficiency comparison}. Bubble size indicates memory footprint. Lower-right region is optimal; (c) Attention map from Qwen2.5-VL-7B \cite{bai2025qwen2} on an input comprising video tokens sampled from four frames in LongVideoBench \cite{wu2024longvideobench} and a language prompt; (d) Long-tailed distribution of attention scores.}
		\label{fig:teaser}
	\end{center}
}]

\begin{abstract}
Long video understanding is essential for human-like intelligence, enabling coherent perception and reasoning over extended temporal contexts. While the emerging thinking-with-frames paradigm—which alternates between global temporal reasoning and local frame examination—has advanced the reasoning capabilities of video multi-modal large language models (MLLMs), it suffers from a significant efficiency bottleneck due to the progressively growing and redundant multi-modal context. To address this, we propose SpecTemp, a reinforcement learning-based Speculative Temporal reasoning framework that decouples temporal perception from reasoning via a cooperative dual-model design. In SpecTemp, a lightweight draft MLLM rapidly explores and proposes salient frames from densely sampled temporal regions, while a powerful target MLLM focuses on temporal reasoning and verifies the draft’s proposals, iteratively refining its attention until convergence. This design mirrors the collaborative pathways of the human brain, balancing efficiency with accuracy. To support training, we construct the SpecTemp-80K dataset, featuring synchronized dual-level annotations for coarse evidence spans and fine-grained frame-level evidence. Experiments across multiple video understanding benchmarks demonstrate that SpecTemp not only maintains competitive accuracy but also significantly accelerates inference compared with existing thinking-with-frames methods.
\end{abstract}
\section{Introduction} \label{sec:intro}



Long video understanding constitutes a fundamental component of human-like intelligence, enabling coherent perception, structured memory, and causal reasoning over extended temporal contexts. Recent advances in reinforcement learning with verifiable rewards (RLVR) \cite{guo2025deepseek,shen2025vlm,yang2025r1,peng2025lmm,tan2025reason,liu2025visual,meng2025mm,yu2025perception} have substantially strengthened the reasoning ability of multi-modal large language models (MLLMs) \cite{gpt4o,anthropic2024claude,bai2025qwen2}. Preliminary studies have pioneered the adaptation of RLVR to video understanding with task-specific reward formulations \cite{feng2025video,park2025deepvideo,li2025videochat,chen2025scaling,wang2025time,tang2025video}. To better accommodate complex spatio-temporal understanding, recent studies have proposed the \emph{thinking-with-frames} \cite{yan2025videochat,he2025framethinker,fu2025love,li2025tempsamp,tang2025tspo,yao2025generative,zhang2025thinking} paradigm. As shown in Figure \ref{fig:teaser}(a), instead of processing an entire video sequence in a single pass, this paradigm alternates between two complementary reasoning pathways: extracting temporal clues from long-horizon segments to identify salient or informative temporal regions, and examining densely sampled frames within those regions to refine local understanding. Through iterative refinement, video MLLMs progressively concentrate on informative segments, facilitating temporally grounded reasoning over extended video contexts.


Despite its promise, the current thinking-with-frames paradigm suffers from an efficiency bottleneck induced by long multi-modal contexts. Existing methods progressively maintain extensive multi-modal sequences that include both high-level reasoning traces and densely sampled visual tokens, which substantially hinders inference \emph{efficiency} (\cf Figure \ref{fig:teaser}(b)). As illustrated in Figure \ref{fig:teaser}(c), we observe a clear modality boundary where language tokens selectively attend to only a small subset of video tokens, revealing the high visual redundancy. Moreover, the excessively long context exhibits significant \emph{redundancy}. For instance in Figure \ref{fig:teaser}(d), we analyze the attention scores over the visual tokens and observe a clear long-tailed distribution, where a considerable portion of visual tokens receive extremely low attention weights. Specifically, more than 90\% of the visual tokens have attention scores below $10^{-3}$, indicating that these tokens contribute minimally to reasoning and answer generation.

A recent research direction for accelerating MLLM decoding is speculative decoding \cite{leviathan2023fast,chen2023accelerating,gagrani2024speculative}, which leverages a lightweight draft model to \emph{speculate} intermediate reasoning tokens and a stronger target model to \emph{verify} the generated tokens. This paradigm allows significant acceleration while maintaining or even improving reasoning accuracy. Inspired by these insights, we propose a reinforcement learning based \textbf{Spec}ulative \textbf{Temp}oral Reasoning (\textbf{SpecTemp}) framework to decouple temporal perception from reasoning through a cooperative dual-model design. As shown in Figure \ref{fig:teaser}(a), SpecTemp delegates the frame selection process to a lightweight draft MLLM, while reserving the target MLLM for temporal reasoning and verification of the draft's proposals. In terms of workflow, we first uniformly sample video frames and feed them into the target MLLM, which performs initial temporal reasoning and predicts temporal clues that require further attention. A lightweight draft model then conducts dense sampling within the predicted temporal regions to explore fine-grained temporal cues, while producing a compact set of sparse representative frames returned to the target model. The target model then verifies the proposed frames, decides whether sufficient information has been gathered, or iteratively triggers another round of draft speculation and verification until convergence. Through this dual-system design, our SpecTemp design mirrors human brain's cortical collaborations \cite{somervail2025two}, where a fast perceptual subsystem (akin to the \emph{lemniscal} pathway) rapidly explores the scene, and a slower cognitive module (\ie, the \emph{extralemniscal} pathway) validates and integrates insights.

To facilitate the training of SpecTemp, we construct SpecTemp-80K, the large-scale datasets tailored for dual-model collaborative reasoning. Each sample contains synchronized dual-level annotations, \ie, coarse answer-relevant evidence spans for the target model, and fine-grained frame-level evidence for the draft model. Through consecutive supervised fine-tuning and reinforcement learning on the SpecTemp-80K dataset, SpecTemp demonstrates consistent improvements across 8 video understanding benchmarks. In addition, compared to prior works following the thinking-with-frames paradigm, our SpecTemp exhibits a clear advantage in efficiency.

In summary, our key contributions are as follows:
\begin{itemize}[topsep=0pt, partopsep=0pt, leftmargin=13pt, parsep=0pt, itemsep=3pt]
    \item  Speculative Visual Reasoning (SpecTemp): We propose a hierarchical test-time scaling framework that integrates speculative reasoning into iterative video perception, enabling semantic-level approximation and verification between models of different capacities.

    \item Collaborative Sampling Mechanism: We introduce a cooperative dense–sparse frame sampling strategy, where the small model speculates local visual details while the base model dynamically validates and refines global spatio-temporal focus.

    \item  Empirical Validation: Extensive experiments on video reasoning benchmarks demonstrate that SpecTemp achieves comparable or superior accuracy to VideoChat-R1.5 \cite{yan2025videochat}, while reducing inference latency by approximately 20\%, highlighting its potential for real-time multimodal reasoning.    
\end{itemize}
\section{Related Work} \label{sec:related}
\subsection{Long Video Understanding}
Although MLLMs have made considerable advances in vision-language tasks, they still struggle with long-form videos exceeding minute-level durations, due to complex dynamics and limited context \cite{zou2024seconds,tang2025video}. Early approaches primarily focused on token compression \cite{li2024llama,song2024moviechat,shen2024longvu,man2025adacm,shu2025video} or keyframe selection \cite{zhang2025q,wang2025videoitg,hu2025m}, yet such methods often sacrifice fine-grained details due to the high compression ratio. Another line of work \nocite{cao2023iterative,yang2021rr,zhou2023exploring,cao2022correspondence,cao2021all,cao2025ground,xu2025a0,ma2025scalelong,chen2025transmamba} explores extending the LLM's context window \cite{chen2024longvila,shen2025long,ren2025vamba}, though this does not fundamentally alleviate the high computational burden and processing costs associated with long videos. Agent-based methods \cite{luo2024video,liu2025videomind,wang2025videochat,chen2025lvagent}, on the other hand, typically divide long videos into sub-segments and leverage external tools to gather additional information. However, these approaches often require laborious customization and involve complex post-processing. As a result, an effective and scalable framework for long-form video understanding remains elusive.


\subsection{Thinking with frames}

With the recent success of reinforcement learning in enhancing the reasoning capabilities of MLLMs \cite{guo2025deepseek,ouyang2022training,jaech2024openai}, the \emph{thinking-with-frames} paradigm \cite{xie2025video,zhang2025thinking,wang2025video,he2025framethinker,fu2025love,yan2025videochat,tang2025tspo,yao2025k,cao2025ground} has attracted growing attention. This approach adaptively zooms into specific video clips and performs interleaved reasoning over textual and visual contexts in a unified manner, enabling more focused reasoning and improved temporal grounding. However, despite its effectiveness, it still suffers from efficiency issues: the sampled tokens from predicted temporal cues remain numerous, and the resulting visual representations are often highly redundant. To address these limitations, our proposed SpecTemp decouples temporal perception from video reasoning and employs a lightweight MLLM to extract the most relevant sparse frames,\nocite{cao2021pursuit,zhang2021cola,cao2022locvtp,zhang2022unsupervised,liu2023qilin,li2023g2l,cao2022deep,ye2023qilin,cao2021unifacegan} thereby achieving significant efficiency gains without sacrificing reasoning quality.

\subsection{Speculative Decoding}

Speculative decoding \cite{leviathan2023fast,xia2024unlocking,chen2023accelerating} has demonstrated its effectiveness in accelerating large language models by employing a smaller draft model to rapidly generate multiple tokens that are then verified in parallel by the target LLM. Most existing approaches focus on token-level acceleration. Early works \cite{cai2024medusa,li2024eagle,du2024glide} trained lightweight draft models to achieve high acceleration ratios through reduced latency. Self-speculative methods \cite{xia2024swift,zhang2023draft} further improve efficiency by leveraging the original model's internal information without introducing auxiliary models. With the recent emphasis on reasoning capabilities in LLMs \cite{guo2025deepseek,ouyang2022training,jaech2024openai}, several studies \cite{yang2025speculative,pan2025specreason,wang2025accelerating,fu2025scaling} have shifted toward reasoning-level speculative acceleration, which differs fundamentally from traditional token-level speculative decoding. Speculative Thinking \cite{yang2025speculative} follows this paradigm by allowing a small draft model to generate most of the reasoning process while selectively delegating challenging reasoning segments to a stronger target model. Inspired by this principle, we extend the reasoning-level speculative strategy to long video understanding, where the time-consuming frame zoom-in process in the original thinking-with-images paradigm is delegated to a small draft model, yielding significant speedups.






\section{Methodology} \label{sec:method}

\begin{figure*}[t]
	\centering
        \includegraphics[width=0.98\textwidth]{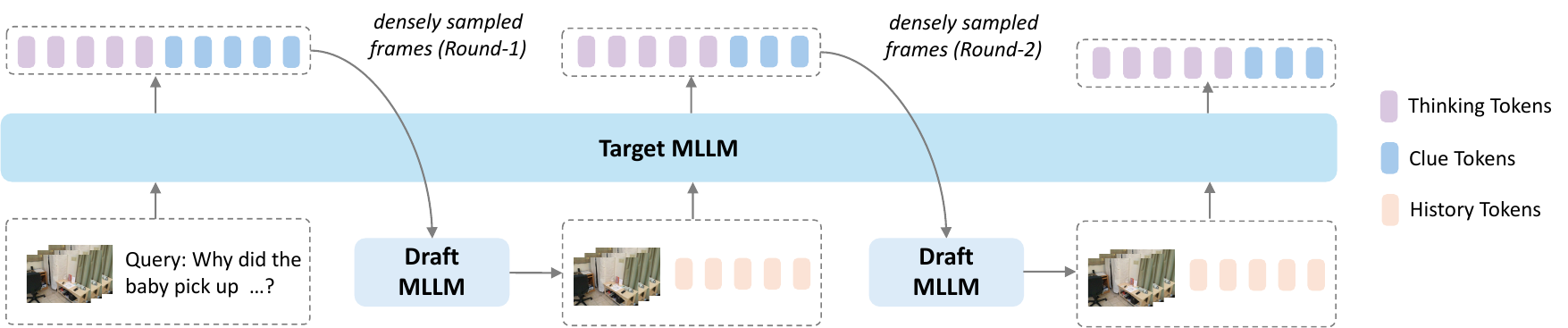}
     \caption{\textbf{Illustration of SpecTemp's iterative speculation-verification process}. Target MLLM predicts temporal regions (Clue Tokens); Draft MLLM proposes frames; Target MLLM verifies with History Tokens across iterative rounds.}
	\label{fig:pipeline}
\end{figure*}

\subsection{Overview}

Our SpecTemp introduces a dual-system cooperative framework that enables efficient reasoning and resampling over long videos. We further formalize both the vanilla MLLMs and our proposed SpecTemp for conceptual comparison.

\noindent \textbf{Vanilla MLLMs:} Given the video $\mathbf{V}$ and a textual query $\mathbf{W}^{\mathrm{q}}$, the vanilla MLLMs $\pi(\cdot)$ formulates the conditional likelihood of generating the response $\mathbf{W}^a$ as follows:
\begin{equation}
\pi\left(\mathbf{W}^{\mathrm{a}} \mid \mathbf{W}^{\mathrm{q}}, \mathbf{V}\right)=\prod_{i=1}^n \pi\left(\mathbf{W}^{\mathrm{a}}_i \mid \mathbf{W}^{\mathrm{a}}_{<i}, \mathbf{W}^{\mathrm{q}}, \mathbf{V}\right),
\end{equation}
where $n$ denotes the length of the answer sequence $\mathbf{W}^a$.

\noindent \textbf{SpecTemp:} Our proposed SpecTemp is composed of a target model $\pi_{\text{target}}$ responsible for temporal reasoning and verification as well as a draft model $\pi_{\text{draft}}$ specialized in dense perception and fine-grained frame selection. The likelihood of the textual response $\mathbf{W}^a$ is estimated as follows:
\begin{equation}
\prod\nolimits_{\substack{\le T_{\text{max}}}} \pi_{\text{target}}\!\left(\mathbf{W}^{\mathrm{a}}, \mathbf{V}^{\mathrm{d}} \mid \mathbf{W}^{\mathrm{q}}, \mathbf{V}^{\mathrm{s}}\right)
\cdot
\pi_{\text{draft}}\!\left(\mathbf{V}^{\mathrm{s}} \mid \mathbf{V}^{\mathrm{d}} \right),
\end{equation}
\noindent where $\mathbf{V}^{\mathrm{d}}$ denotes densely sampled frames (\eg, 1 fps) from the attentive regions predicted by the target MLLM, and $\mathbf{V}^{\mathrm{s}}$ represents the salient frames (\eg, 2 frames) selected by the lightweight draft MLLM as the most informative subset. $T_{\text{max}}$ denotes the maximum iteration number. The computational asymmetry satisfies $|\pi_{\text{draft}}| \ll |\pi_{\text{target}}|$, where $|\pi_{\text{draft}}|$ and $|\pi_{\text{target}}|$ denote the number of trainable parameters of the draft and target models, respectively. This asymmetric design allows the draft model to perform rapid dense perception with minimal computational cost, while the target model focuses on high-level temporal reasoning and semantic verification.

The overall algorithm is illustrated in Algorithm \ref{alg:algo} and visualized in 
Figure \ref{fig:pipeline}, showing the iterative speculation-verification process. We sequentially describe the processes of initialization, speculative exploration by the draft MLLM, and verification by the target MLLM.

\begin{algorithm}[t]
\caption{Speculative Temporal Reasoning (SpecTemp)}
\label{alg:algo}
\begin{algorithmic}[1]
\Require Long video $\mathbf{V}$, query $\mathbf{W}^{\mathrm{q}}$
\Ensure Answer $a$
\State $\mathcal{F}^{(0)} \leftarrow \text{UniformSample}(\mathbf{V}, K)$
\State $\mathbf{T}_{0}, \mathcal{R}_{0}, \mathbf{W}^{\mathrm{a}} \leftarrow \mathcal{M}_T(\mathcal{F}_{0}, \mathbf{W}^{\mathrm{q}})$ \Comment{$\mathbf{T}$: reasoning trace, $\mathcal{R}$: temporal RoI, $\mathbf{W}^{\mathrm{a}}$: answer}
\If{$\mathbf{W}^{\mathrm{a}} \neq \emptyset$}
    \State \Return $\mathbf{W}^{\mathrm{a}}$ \Comment{Early termination if answer is already obtained}
\EndIf
\State $t \leftarrow 1$
\While{$t \leq T_\text{max}$}
    \State $\mathcal{F}^\text{dense}_{t} \leftarrow \text{DenseSample}(\mathbf{V}, \mathcal{R}_{t-1}, \delta)$\Comment{$\delta$: Sampling fps}
    \State $\mathcal{F}^\text{draft}_{t} \leftarrow \mathcal{M}_D(\mathcal{F}^\text{dense}_{t} \mid \mathbf{T}_{t-1})$ \Comment{Draft uses history context only from the current iteration}
    \State $\mathbf{T}_{t}, \mathcal{R}_{t}, \mathbf{W}^{\mathrm{a}} \leftarrow \mathcal{M}_T(\mathcal{F}^\text{draft}_{t} \mid \mathbf{W}^{\mathrm{q}},  \mathbf{T}_{<t})$ \Comment{Target uses all previous history context}
    \If{$\mathbf{W}^{\mathrm{a}} \neq \emptyset$}
        \State \Return $\mathbf{W}^{\mathrm{a}}$ \Comment{Return answer if obtained during iteration}
    \EndIf
    \State $t \leftarrow t + 1$
\EndWhile
\State $\mathbf{W}^{\mathrm{a}} \leftarrow \mathcal{M}_T.\text{generate\_answer}(\mathcal{F}_{t}, \mathbf{W}^{\mathrm{q}})$ \Comment{Final answer generation if not converged earlier}
\State \Return $\mathbf{W}^{\mathrm{a}}$
\end{algorithmic}
\end{algorithm}


\subsection{Speculative Temporal Reasoning}

\noindent \textbf{Initialization.} Given the input video $\mathbf{V}$, we first feed the uniformly sampled frames $\mathbf{V}^{\mathrm{s}}_{0}$, together with the query $\mathbf{W}^q$, into target MLLMs for initial reasoning.
\begin{equation}
\mathbf{T}_{0}, \mathbf{V}^{\mathrm{d}}_{0}, \mathbf{W}^\mathrm{a} = \pi_{\text{target}}\left(\mathbf{W}^\mathrm{q}, \mathbf{V}^{\mathrm{s}}_{0}\right)
\end{equation}
\noindent where $\mathbf{T}_{0}$ represents the initial reasoning trace, $\mathbf{V}^{\mathrm{d}}_{0}$ denotes the densely sampled frames from the predicted temporal region which requires further examination. $\mathbf{W}^\mathrm{a}$ is the final answer if immediately determinable. 

Notably, $\mathbf{T}_{0}$ and $\mathbf{W}^\mathrm{a}$ are \emph{mutually exclusive}. That is, when the target model $\mathcal{M}_T$ determines that sufficient temporal evidence is available, it directly outputs the answer; otherwise, it predicts a temporal evidence region to trigger further dense sampling.

\noindent \textbf{Speculation} Given the densely sampled frames $\mathbf{V}^{\mathrm{d}}_t$ from the target MLLM, the draft MLLM conditions on the reasoning trace $\mathbf{T}_{t-1}$ of the current iteration to determine the most informative sparse video frames $\mathbf{V}^{\mathrm{s}}_t$:
\begin{equation}
\mathbf{V}^{\mathrm{s}}_t  = \pi_{\text{draft}}\left(\mathbf{V}^{\mathrm{d}}_t  ; \mathrm{T}_{t-1}\right), \quad t \in [1, T_{\text{max}}].
\end{equation}

\noindent \textbf{Verification.} The target model reasons and verifies the speculative proposal from the draft model:
\begin{equation}
(\mathbf{T}_{t}, \mathbf{V}^{\mathrm{d}}_{t}, \mathbf{W}^\mathrm{a}) = \pi_{\text{target}}\!\left( \mathbf{V}^{\mathrm{s}}_t  \,;\,  \mathbf{T}_{<t} \right),
\end{equation}
where $\mathbf{T}_{<t} = \{\mathbf{T}_{0}, \mathbf{T}_{1}, \ldots, \mathbf{T}_{t-1}\}$ denotes the accumulated reasoning trace across all previous rounds. 

\begin{figure*}[!t]
    \centering
        \includegraphics[width=0.98\textwidth]{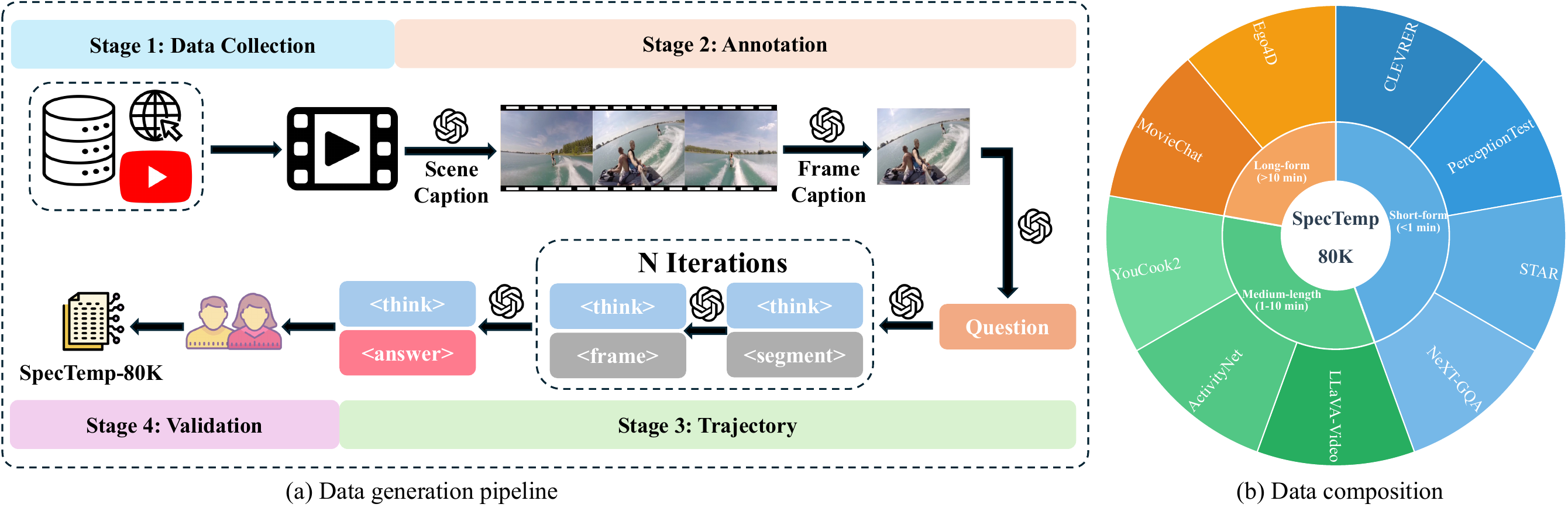}
     \caption{Overview of data generation pipeline and composition}
    \label{fig:data_overview}
\end{figure*}

\subsection{Policy Optimization}

\noindent \textbf{Optimization.} Our SpecTemp undergoes a two-stage optimization process including cold-start supervised fine-tuning (SFT) and reinforcement fine-tuning (RFT). For SFT, we separately build the high-level verification and fine-grained frame-level reasoning capabilities of the target and draft MLLMs. For RFT, we follow GRPO \cite{shao2024deepseekmath} to model keyframe speculation and response generation as a unified policy learning process:
\begin{equation}
\small
\label{eq:grpo_objective}
\begin{aligned}
& \mathcal{J}=\mathbb{E}\left[\mathbf{W}^{\mathrm{q}}, \mathbf{V},\{\mathbf{W}^{\mathrm{a}}, \mathbf{V}^{\mathrm{d}}\} \sim\pi_{\text{target}_{\text{old}}}, \mathbf{V}^{\mathrm{s}} \sim \pi_{\text{draft}_{\text{old}}} \right] \\
& \frac{1}{G} \sum_{i=1}^G \frac{\pi_{\text{target}}\!\left(\mathbf{W}^{\mathrm{a}}_i, \mathbf{V}^{\mathrm{d}} \mid \mathbf{W}^{\mathrm{q}}, \mathbf{V}^{\mathrm{s}}\right)
\cdot
\pi_{\text{draft}}\!\left(\mathbf{V}^{\mathrm{s}} \mid \mathbf{V}^{\mathrm{d}} \right)}{\pi_{\text{target}_\text{old}}\!\left(\mathbf{W}^{\mathrm{a}}_i, \mathbf{V}^{\mathrm{d}} \mid \mathbf{W}^{\mathrm{q}}, \mathbf{V}^{\mathrm{s}}\right)
\cdot
\pi_{\text{draft}_\text{old}}\!\left(\mathbf{V}^{\mathrm{s}} \mid \mathbf{V}^{\mathrm{d}} \right)} A_i \\
& \quad-\beta \cdot \mathbb{D}_{K L}\left(\pi_{\text{target}} \cdot \pi_{\text{draft}} \| \pi_{\text{target}_{\text{ref}}} \cdot \pi_{\text{draft}_{\text{ref}}}\right),
\end{aligned}
\end{equation}
where $A_i$ denotes the advantage computed following group rewards. Note that the clipping operation in Eq \eqref{eq:grpo_objective} is omitted for clarity.


\noindent\textbf{Reward Design.}
Our training objective employs distinct reward formulations for the target and draft models to encourage their specialized behaviors in the collaborative framework.

\noindent\textit{Target Model Rewards.}
The target model receives three complementary reward signals: (1) \textbf{Format reward} $R_{\text{format}}^{\text{target}}$ evaluates whether the output follows the prescribed structure with proper \texttt{<think>}, \texttt{<segment>}, and \texttt{<answer>} tokens; (2) \textbf{Answer reward} $R_{\text{answer}}$ measures the correctness of the final response by comparing against ground-truth annotations; and (3) \textbf{IoU reward} $R_{\text{IoU}}$ assesses the temporal localization quality by computing the Intersection-over-Union between predicted evidence segments and ground-truth temporal regions. The total reward for the target model is:
\begin{equation}
R_{\text{target}} = R_{\text{format}}^{\text{target}} + R_{\text{answer}} + R_{\text{IoU}}.
\end{equation}

\noindent\textit{Draft Model Rewards.}
The draft model is optimized with two reward components: (1) \textbf{Format reward} $R_{\text{format}}^{\text{draft}}$ ensures proper frame selection output format; and (2) \textbf{Visual information gain reward} $R_{\text{visual}}$ encourages the selection of informative and diverse frames. The visual information gain reward consists of two terms computed via CLIP-based similarity scores:
\begin{equation}
R_{\text{visual}} = \text{Sim}_{\text{CLIP}}(q, f_i) - \max_{f_j \in \mathcal{F}_{\text{prev}}} \text{Sim}_{\text{CLIP}}(f_i, f_j),
\end{equation}
where $q$ denotes the question text, $f_i$ represents the selected frame, and $\mathcal{F}_{\text{prev}}$ is the set of previously selected frames. The first term rewards frames with high relevance to the question, while the second term penalizes redundancy by discouraging selection of frames similar to existing ones. The total draft model reward is:
\begin{equation}
R_{\text{draft}} = R_{\text{format}}^{\text{draft}} + R_{\text{visual}}.
\end{equation}


\subsection{SpecTemp-80K Dataset Construction}

To facilitate training of our dual-model framework, we construct \textbf{SpecTemp-80K}, a large-scale dataset with synchronized dual-level annotations. As illustrated in Figure~\ref{fig:data_overview}, our pipeline consists of four stages: data collection, annotation, trajectory generation, and validation.

\noindent\textbf{Data Collection.}
We curate videos from diverse sources spanning three duration categories: (1) \textit{Short-form} ($<$1 min) from CLEVRER, PerceptionTest, STAR, and NeXT-GQA; (2) \textit{Medium-length} (1-10 min) from LLaVA-Video, ActivityNet, and YouCook2; and (3) \textit{Long-form} ($>$10 min) from MovieChat and Ego4D.

\noindent\textbf{Annotation.}
We leverage GPT-4o~\cite{gpt4o} to generate hierarchical annotations for dual-level supervision. For each video, GPT-4o produces \textit{scene-level captions} describing overall narratives and key events, as well as \textit{frame-level captions} capturing fine-grained visual details. These annotations provide synchronized supervision for both models: coarse temporal evidence spans for the target model's reasoning and fine-grained frame evidence for the draft model's selection.

\noindent\textbf{Trajectory Generation.}
Using GPT-4o, we synthesize multi-round reasoning trajectories that simulate SpecTemp's iterative speculation-verification process. Given the video, question, and dual-level annotations, GPT-4o generates structured trajectories with \texttt{<think>} tokens for temporal reasoning, \texttt{<segment>} tokens indicating evidence regions that require dense sampling, \texttt{<frame>} tokens for the draft model's selected salient frames, and \texttt{<answer>} tokens for final responses. This automated generation enables scalable creation of high-quality training data.

\noindent\textbf{Validation.}
We manually verify a subset of generated trajectories to ensure annotation quality and reasoning consistency, filtering out trajectories with incorrect temporal evidence or inconsistent frame selections.



\section{Experiments} \label{sec:experiments}

\subsection{Experimental Setup}

\begin{table*}[!ht]
\centering
\caption{Performance of different models on short-form video benchmarks.}
\label{tab:main_results_short_form}
\resizebox{0.98\textwidth}{!}{
\fontsize{9}{11}\selectfont
\begin{tabular}{lcccccc}
\toprule
\textbf{Models} & \textbf{Frames} & \textbf{TempCompass \cite{liu2024tempcompass}} & \textbf{MVBench \cite{li2024mvbench}} & \textbf{MMVU(mc) \cite{zhao2025mmvu}} & \textbf{VSI-Bench \cite{yang2025thinking}} & \textbf{Latency(s)} \\
Average duration (s) &  & 12 & 19 & 52 & 117 & \\
\hline
\multicolumn{6}{c}{\emph{\textbf{Proprietary Multi-modal LLMs}}}  \\ 
GPT‑4o \cite{gpt4o} & 16 & 71.1 & 52.6 & 71.5 & 42.6 & - \\
Gemini 2.5 Pro \cite{comanici2025gemini} & 16 & 77.6 & 58.0 & 75.7 & 45.1 & - \\
\hline
\multicolumn{6}{c}{\emph{\textbf{Open-source Multi-modal LLMs}}}  \\
LongVA-7B \cite{zhang2024long} & 16 & 73.8 & 66.4 & 66.1 & 35.9 & 2.7 \\
InternVL3-8B \cite{zhu2025internvl3} & 16 & 72.1 & 65.1 & 66.2 & 36.4 & 2.1 \\
Video-R1 \cite{feng2025video} & 16 & 71.5 & 62.7 & 64.0 & 29.8 & 2.4 \\
InternVL3-8B \cite{zhu2025internvl3} & 64 & 75.4 & 68.3 & 68.2 & 39.1 & 5.0 \\
VideoChat-Flash \cite{li2024videochat} & 64 & 74.4 & 66.8 & 67.2 & 37.4 & 4.9 \\
VideoChat-R1.5 \cite{yan2025videochat} & 64 & 77.3 & 70.6 & 70.1 & 39.4 & 5.8 \\
\hline
Qwen2.5-VL-7B \cite{bai2025qwen2} & 16 & 72.2 & 64.1 & 65.7 & 34.3 & 2.1 \\
\rowcolor{gray!15}
\textbf{SpecTemp(Ours)} & 13.7 & 75.3$_{+3.1}$ & 68.7$_{+4.6}$ & 67.8$_{+2.1}$ & 37.4$_{+3.1}$ & 1.8 \\
Qwen2.5-VL-7B \cite{bai2025qwen2} & 64 & 74.2 & 66.5 & 67.8 & 36.6 & 5.1 \\
\rowcolor{gray!15}
\textbf{SpecTemp(Ours)} & 47.6 & 77.2$_{+3.0}$ & 69.3$_{+2.8}$ & 70.4$_{+2.6}$ & 39.7$_{+3.1}$ & 4.7 \\
\bottomrule
\end{tabular}
}
\end{table*}

\begin{table*}[!ht]
\centering
\caption{Performance of different models on long-form video benchmarks.}
\label{tab:main_results_long_form}
\resizebox{0.98\textwidth}{!}{
\fontsize{13}{15}\selectfont
\begin{tabular}{lcccccc}
\toprule
\textbf{Models} & \textbf{Frames} & \textbf{Video-Holmes \cite{cheng2025video}} & \textbf{LongVideoBench \cite{wu2024longvideobench}} & \textbf{MLVU \cite{zhou2024mlvu}} & \textbf{Video-MME(wo sub) \cite{fu2025video}} & \textbf{Latency(s)} \\
Average duration (s) &  & 165 & 732 & 933 & 1021 \\
\hline
\multicolumn{6}{c}{\emph{\textbf{Proprietary Multi-modal LLMs}}}  \\ 
GPT‑4o \cite{gpt4o} & 16 & 44.4 & 50.6 & 43.8 & 62.0 & - \\
Gemini 2.5 Pro \cite{comanici2025gemini} & 16 & 43.8 & 51.5 & 44.2 & 65.6 & - \\
\hline
\multicolumn{6}{c}{\emph{\textbf{Open-source Multi-modal LLMs}}}  \\
LongVA-7B \cite{zhang2024long} & 16 & 43.2 & 56.8 & 47.5 & 60.1 & 4.1 \\
InternVL3-8B \cite{zhu2025internvl3} & 16 & 40.1 & 56.1 & 49.0 & 61.8 & 3.9 \\
Video-R1 \cite{feng2025video} & 16 & 39.8 & 54.6 & 45.6 & 57.4 & 4.7 \\
InternVL3-8B \cite{zhu2025internvl3} & 64 & 44.0 & 60.2 & 51.4 & 62.2 & 9.1 \\
VideoChat-Flash \cite{li2024videochat} & 64 & 43.5 & 59.7 & 51.5 & 63.0 & 10.7 \\
VideoChat-R1.5 \cite{yan2025videochat} & 64 & 45.1 & 60.6 & 52.3 & 63.4 & 11.5 \\
\hline
Qwen2.5-VL-7B \cite{bai2025qwen2} & 16 & 35.0 & 54.5 & 40.6 & 56.0 & 4.1 \\
\rowcolor{gray!15}
\textbf{SpecTemp(Ours)} & 14.5 & 47.0$_{+12.0}$ & 57.5$_{+3.0}$ & 48.6$_{+8.0}$ & 62.4$_{+6.4}$ & 3.7 \\
Qwen2.5-VL-7B \cite{bai2025qwen2} & 64 & 37.8 & 57.6 & 48.2 & 61.4 & 9.7 \\
\rowcolor{gray!15}
\textbf{SpecTemp(Ours)} & 58.1 & 47.8$_{+10.0}$ & 61.4$_{+3.8}$ & 50.9$_{+2.7}$ & 64.1$_{+2.7}$ & 8.9 \\
\bottomrule
\end{tabular}
}
\end{table*}

\noindent\textbf{Implementation Details.} 
Our SpecTemp framework consists of a 7B-parameter target MLLM and a 3B-parameter draft MLLM, both initialized from Qwen2.5-VL \cite{bai2025qwen2}. For the target model, we uniformly sample 10 frames during initialization. The draft model performs dense sampling at 1 fps within predicted temporal regions and selects 2 representative frames per iteration. We set the maximum iteration number to 3 to balance performance and efficiency.

\noindent\textbf{Benchmarks.} 
We conduct comprehensive evaluations across eight widely-used video understanding benchmarks, spanning both short-form and long-form scenarios. For short-form videos, we evaluate on: (1) \textbf{TempCompass} \cite{liu2024tempcompass}, which assesses fine-grained temporal perception through conflicting video pairs that differ only in specific temporal aspects; (2) \textbf{MVBench} \cite{li2024mvbench}, featuring 20 challenging tasks covering temporal skills from perception to cognition via a static-to-dynamic design; (3) \textbf{MMVU} \cite{zhao2025mmvu}, requiring expert-level reasoning with domain-specific knowledge across 27 subjects in four core disciplines; and (4) \textbf{VSI-Bench} \cite{yang2025thinking}, evaluating visual-spatial intelligence through tasks that require tracking spatial relationships across sequences. For long-form videos, we use: (1) \textbf{Video-Holmes} \cite{cheng2025video}, requiring models to actively locate and connect multiple scattered clues for complex multi-step reasoning; (2) \textbf{LongVideoBench} \cite{wu2024longvideobench}, featuring 6,678 questions with the referring reasoning paradigm for accurate retrieval from video-language interleaved inputs; (3) \textbf{MLVU} \cite{zhou2024mlvu}, encompassing nine diversified tasks across various video genres to assess both global and local understanding; and (4) \textbf{Video-MME} \cite{fu2025video}, distinguished by integrating multiple data modalities (video, subtitles, audio) across 6 visual domains with 30 subfields.


\noindent\textbf{Evaluation Metrics.} 
Following standard practices, we report accuracy for all benchmarks and measure average inference latency per sample under identical hardware configurations (batch size of 1). For fair comparison, SpecTemp is configured with 10 initial frames plus up to 3 iterations of 2 frames each (comparing with 16-frame baselines) and 32 initial frames plus up to 4 iterations of 8 frames each (comparing with 64-frame baselines).

\begin{table*}[!ht]
\centering
\caption{Ablation study on frame selection strategies.}
\label{tab:ablation_frame_selection}
\resizebox{0.98\textwidth}{!}{
\begin{tabular}{l|c|cccc}
\toprule
\textbf{Method} & \textbf{Frames} & \textbf{MMVU(mc) \cite{zhao2025mmvu}} & \textbf{Video-Holmes \cite{cheng2025video}} & \textbf{LongVideoBench \cite{wu2024longvideobench}} & \textbf{Video-MME(wo sub) \cite{fu2025video}} \\
\midrule
Uniform & 16 & 65.8 & 39.5 & 55.3 & 57.3 \\
Target + Uniform & 16 & 66.3 & 43.2 & 55.6 & 58.3 \\
Target + CLIP & 16 & 67.1 & 45.1 & 56.7 & 60.5 \\
\textbf{Ours(Target + Draft)} & \textbf{14.1} & \textbf{67.8} & \textbf{47.0} & \textbf{57.5} & \textbf{62.4} \\
\bottomrule
\end{tabular}
}
\end{table*}

\subsection{Main Results}


\noindent\textbf{Performance on Short-form Videos.} 
Table~\ref{tab:main_results_short_form} presents the performance comparison on short-form video benchmarks. Using only 13.7 frames on average, SpecTemp significantly outperforms Qwen2.5-VL-7B (16 frames) with consistent gains across all benchmarks, while reducing inference latency from 2.1s to 1.8s. When scaling to 47.6 frames, SpecTemp achieves comparable accuracy to VideoChat-R1.5 while being 19\% faster. These results validate that our adaptive frame selection effectively identifies salient temporal information, achieving improved understanding with enhanced computational efficiency.

\noindent\textbf{Performance on Long-form Videos.} 
Table~\ref{tab:main_results_long_form} demonstrates SpecTemp's effectiveness on long-form video understanding tasks. With 14.5 frames on average, SpecTemp achieves substantial improvements over Qwen2.5-VL-7B across all benchmarks, with particularly pronounced gains on Video-Holmes (+12.0\%) and MLVU (+8.0\%), while reducing latency from 4.1s to 3.7s. When scaling to 58.1 frames, SpecTemp outperforms VideoChat-R1.5 while being 23\% faster. These results demonstrate that our dual-model framework effectively handles long-form video understanding challenges, where identifying relevant temporal segments becomes crucial for both accuracy and efficiency.


\subsection{Ablation Studies}

\noindent\textbf{Frame Selection Strategies.} 
Table~\ref{tab:ablation_frame_selection} compares different frame selection approaches across four benchmarks. We evaluate: (1) \textbf{Uniform}: uniform sampling of 16 frames; (2) \textbf{Target + Uniform}: temporal region prediction followed by uniform sampling; (3) \textbf{Target + CLIP}: CLIP-based similarity scores for frame selection; and (4) \textbf{Ours (Target + Draft)}: our dual-model collaborative approach.

Uniform baseline achieves reasonable performance but lacks temporal focus. Target + Uniform improves results by 1.2\%, validating temporal localization. Target + CLIP provides further gains (+2.5\%), but our Target + Draft strategy achieves the best performance across all benchmarks with adaptive frame count (14.1 on average), demonstrating that training a specialized draft model is superior to heuristic-based approaches.

\begin{figure}[!h]
	\centering
        \includegraphics[width=0.49\textwidth]{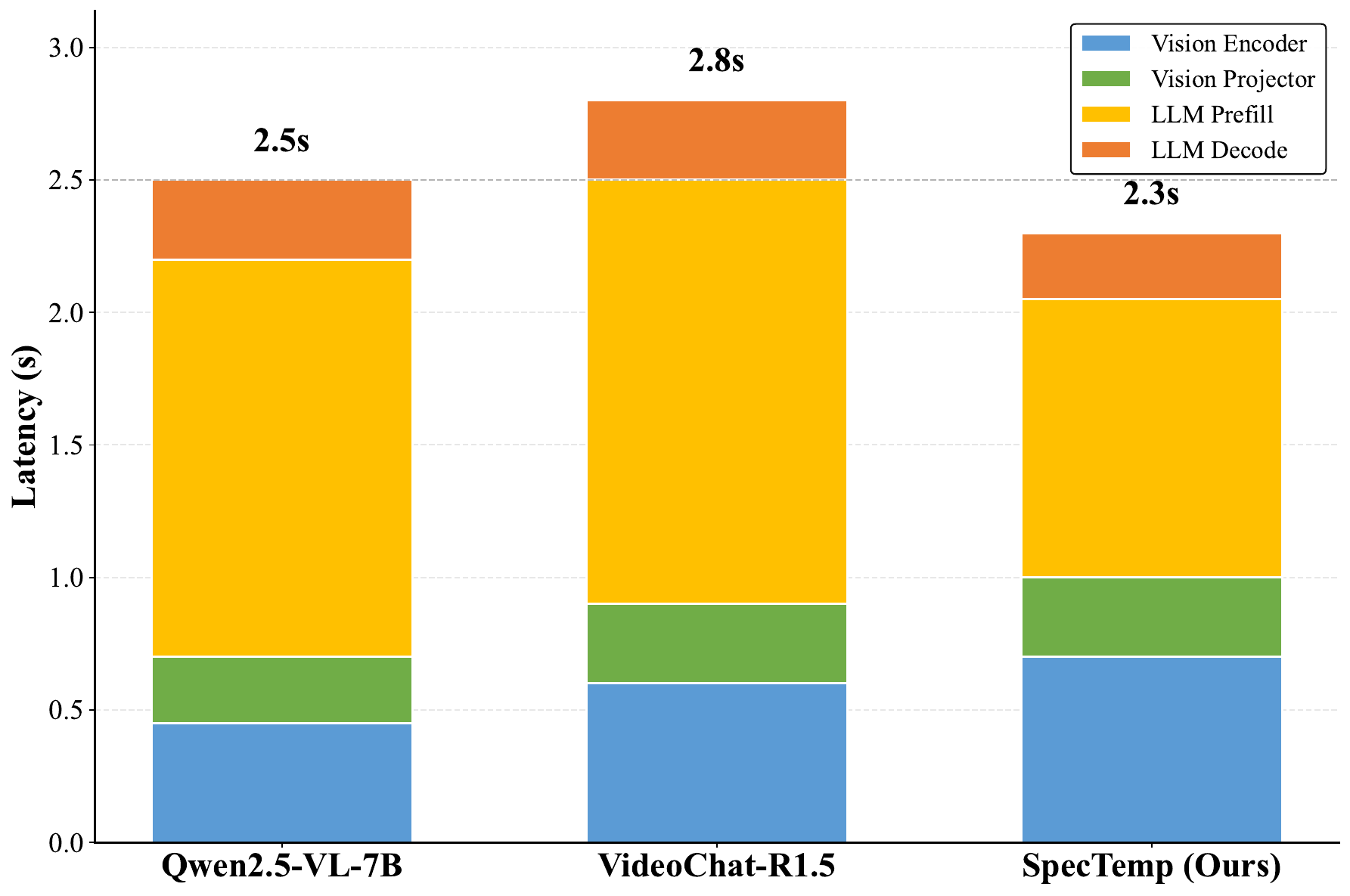}
     \caption{\textbf{Inference latency breakdown}. SpecTemp achieves the lowest latency (2.3s) by delegating dense sampling to the draft model.}
	\label{fig:model_latency_comparison}
\end{figure}

\noindent\textbf{Efficiency Analysis.} 
Figure~\ref{fig:model_latency_comparison} provides a detailed latency breakdown across vision encoder, projector, LLM prefill, and decode stages. The analysis reveals that LLM prefill dominates total inference time. VideoChat-R1.5 incurs the highest latency (2.8s) due to its progressively expanding context with interleaved reasoning and dense visual tokens. SpecTemp achieves the lowest latency (2.3s) by offloading dense exploration to the 3B draft model and selectively proposing only salient frames to the 7B target model. This design minimizes LLM processing overhead—the true bottleneck—while maintaining competitive accuracy, demonstrating the effectiveness of decoupling temporal perception from reasoning.

\begin{table}[h]
\centering
\caption{Ablation study on model collaboration evaluated on LongVideoBench. Efficiency = Accuracy / Latency.}
\label{tab:ablation_model_collaboration}
\resizebox{0.48\textwidth}{!}{
\begin{tabular}{l|cc|ccc}
\toprule
\textbf{Method} & \textbf{Target} & \textbf{Draft} & \textbf{Acc.(\%)} & \textbf{Lat.(s)} & \textbf{Eff.} \\
\midrule
Large only & 7B & 7B & 54.1 & 3.1 & 17.5 \\
Small only & 3B & 3B & 40.3 & \textbf{1.7} & 23.7 \\
\textbf{Ours (L+S)} & \textbf{7B} & \textbf{3B} & \textbf{57.5} & 2.3 & \textbf{25.0} \\
\bottomrule
\end{tabular}
}
\end{table}

\noindent\textbf{Model Collaboration Analysis.} 
Table~\ref{tab:ablation_model_collaboration} analyzes the contribution of model collaboration on LongVideoBench. We compare three configurations: (1) \textbf{Large only (7B)}: using only the 7B model; (2) \textbf{Small only (3B)}: using only the 3B model; and (3) \textbf{Ours (L+S)}: our dual-model collaboration where the 7B target model handles temporal reasoning while the 3B draft model specializes in frame selection.

The results reveal compelling trade-offs between accuracy and efficiency. Large-only achieves high accuracy (54.1\%) but suffers from substantial latency, while Small-only is fast but shows significantly degraded accuracy (40.3\%) due to limited reasoning capacity. Our L+S approach strikes an optimal balance, achieving the highest accuracy (57.5\%) and the best efficiency score (25.0). Compared to Large-only, our method improves both accuracy and latency simultaneously. Compared to Small-only, our approach substantially improves accuracy with only modest latency increase. These results validate our core hypothesis that decoupling temporal perception from high-level reasoning enables both superior performance and computational efficiency.



\begin{table}[!h]
\centering
\caption{Ablation study on training strategies for dual-model collaboration on LongVideoBench.}
\label{tab:training_ablation}
\resizebox{0.48\textwidth}{!}{
\fontsize{8}{8}\selectfont
\begin{tabular}{cc|cc|cc}
\toprule
\multicolumn{2}{c|}{\textbf{Target-7B}} & \multicolumn{2}{c|}{\textbf{Draft-3B}} & \textbf{Acc.} & \textbf{Format Acc.} \\
\cmidrule(lr){1-2} \cmidrule(lr){3-4}
\textbf{SFT} & \textbf{RL} & \textbf{SFT} & \textbf{RL} & \textbf{(\%)} & \textbf{(\%)} \\
\midrule
\multicolumn{6}{l}{\textit{Baseline}} \\
-- & -- & -- & -- & 54.5 & 40.6 \\
\midrule
\multicolumn{6}{l}{\textit{Target Model Only}} \\
\checkmark & -- & -- & -- & 55.3 & 95.2 \\
\checkmark & \checkmark & -- & -- & 56.3 & 96.5 \\
\midrule
\multicolumn{6}{l}{\textit{Dual-Model Collaboration}} \\
\checkmark & -- & \checkmark & -- & 55.8 & 98.7 \\
\checkmark & \checkmark & \checkmark & -- & 57.1 & 99.2 \\
\checkmark & -- & \checkmark & \checkmark & 56.9 & 99.1 \\
\checkmark & \checkmark & \checkmark & \checkmark & \textbf{57.5} & \textbf{99.5} \\
\bottomrule
\end{tabular}
}
\end{table}

\noindent\textbf{Training Strategy Analysis.} 
Table \ref{tab:training_ablation} ablates different training strategies on LongVideoBench. SFT teaches both models the basic output structure and temporal reasoning capabilities, while RL optimizes their collaborative decision-making through reward signals.

Target model SFT improves accuracy to 55.3\% and format accuracy to 95.2\%, establishing proper output structure. Adding RL further boosts accuracy to 56.3\% by optimizing temporal localization through IoU rewards. Introducing draft model SFT significantly improves format accuracy to 98.7\%, as the draft learns structured frame selection. The full strategy (SFT+RL for both models) achieves 57.5\% accuracy and 99.5\% format accuracy. The +3.0\% improvement over baseline validates that SFT provides foundational capabilities while RL refines collaborative reasoning through task-specific optimization.

\begin{figure}[!t]
	\centering
        \includegraphics[width=0.49\textwidth]{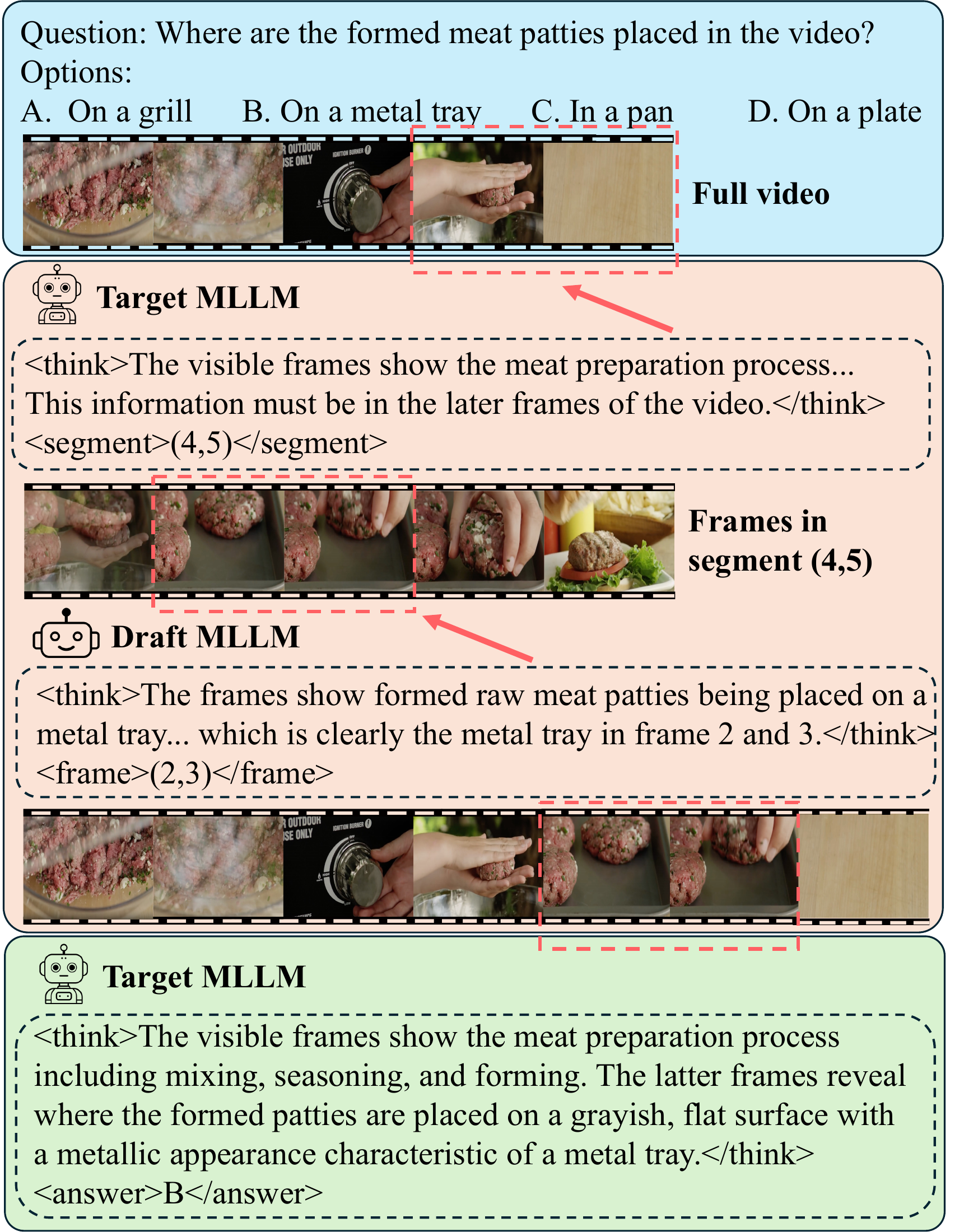}
     \caption{Visualization of SpecTemp's iterative speculation-verification process with target-draft model collaboration.}
	\label{fig:vis_case}
\end{figure}

\subsection{Case Study}

Figure~\ref{fig:vis_case} demonstrates SpecTemp's reasoning process on a cooking video. The target MLLM initially identifies that answering ``Where are the formed meat patties placed at the end of the video?'' requires examining the latter portion, predicting temporal segment (4, 5). The draft MLLM then densely samples this region and proposes key frames showing the patties on a metal tray. The target MLLM verifies these frames and correctly identifies the metal tray as the answer. This example illustrates the effective collaboration between the draft model's rapid perceptual exploration and the target model's careful reasoning, validating our dual-model design for efficient long video understanding.
\section{Conclusions}

In this work, we present \textbf{SpecTemp}, a reinforcement learning-based framework that addresses the efficiency bottleneck in long video understanding through speculative temporal reasoning. By decoupling temporal perception from reasoning via a cooperative dual-model design---where a lightweight 3B draft MLLM explores salient frames and a powerful 7B target MLLM handles temporal reasoning and verification---SpecTemp achieves an optimal balance between accuracy and efficiency. We construct the \textbf{SpecTemp-80K} dataset with dual-level annotations and employ a two-stage optimization protocol. Extensive experiments across eight benchmarks demonstrate that SpecTemp maintains competitive accuracy while accelerating inference by 19-23\% on long-form videos, validating the effectiveness of our speculative temporal reasoning paradigm for efficient video understanding.

\section{Supplementary Material}

\begin{figure*}[!ht]
    \centering
        \includegraphics[width=0.98\textwidth]{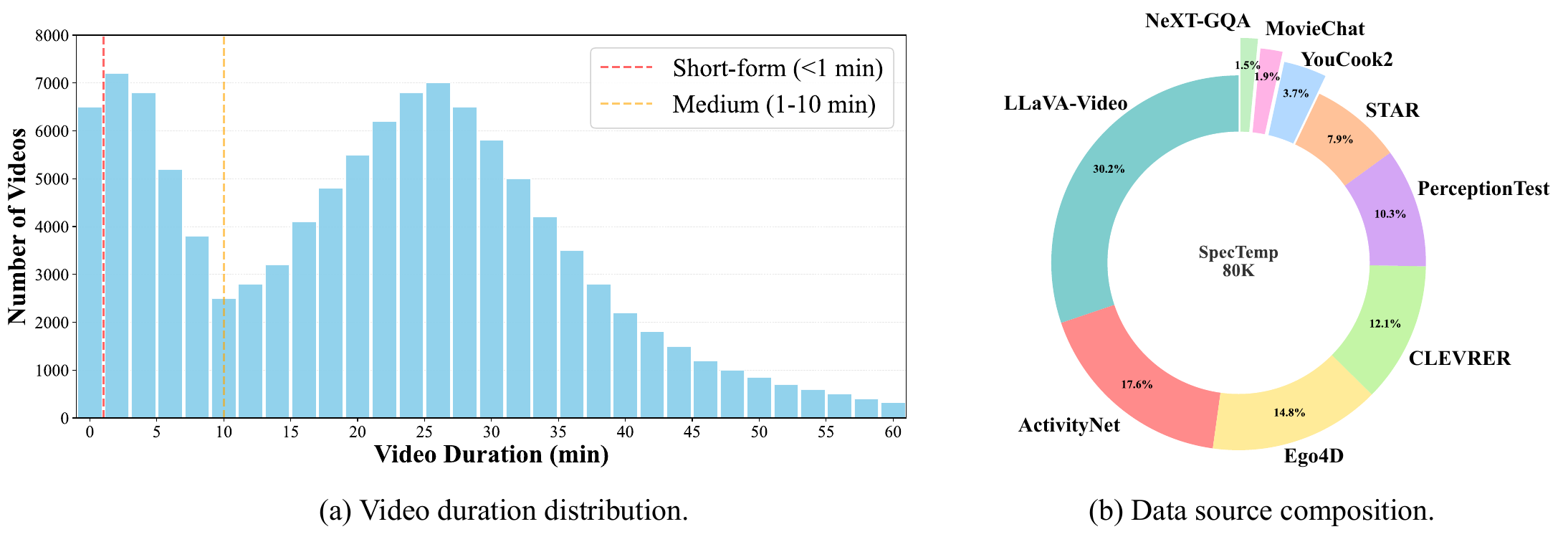}
     \caption{SpecTemp-80K dataset statistics.}
    \label{fig:figDataStatistics_arxiv}
\end{figure*}



This supplementary material is organized as follows. We begin with a detailed description of the experimental setups, then present additional experimental findings, followed by the visualization results and a discussion of limitations.

Specifically, the \textbf{experiment setups} include the following aspects:
\begin{itemize}[topsep=2pt, partopsep=0pt, leftmargin=13pt, parsep=0pt, itemsep=3pt]
    \item Training configurations.
    \item Prompt templates for dual models.
    \item SpecTemp-80K dataset statistics.
\end{itemize}
\textbf{More experimental findings} include the following aspects:
\begin{itemize}[topsep=2pt, partopsep=0pt, leftmargin=13pt, parsep=0pt, itemsep=3pt]
    \item Ablation on different iteration numbers.
    \item Ablation on reward components.
    \item Frame allocation strategies under fixed budget.
    \item Needle-in-a-Haystack evaluation.
\end{itemize}

\subsection{Experiment Setup} 

\noindent \textbf{Training configurations.} We employ a two-stage training strategy. First, we conduct supervised fine-tuning (SFT) for 2 epochs on the SpecTemp-80K dataset with a learning rate of $1\times10^{-6}$. Subsequently, we perform reinforcement fine-tuning (RFT) using GRPO \cite{shao2024deepseekmath} for 1 epoch with a learning rate of $1\times10^{-6}$. The KL penalty coefficient $\beta$ is set to 0.04. All experiments are conducted on 16$\times$H100 GPUs with batch sizes of 32 for SFT and 16 for RFT.

\begin{figure*}[t]
    \centering
    \includegraphics[width=0.92\textwidth]{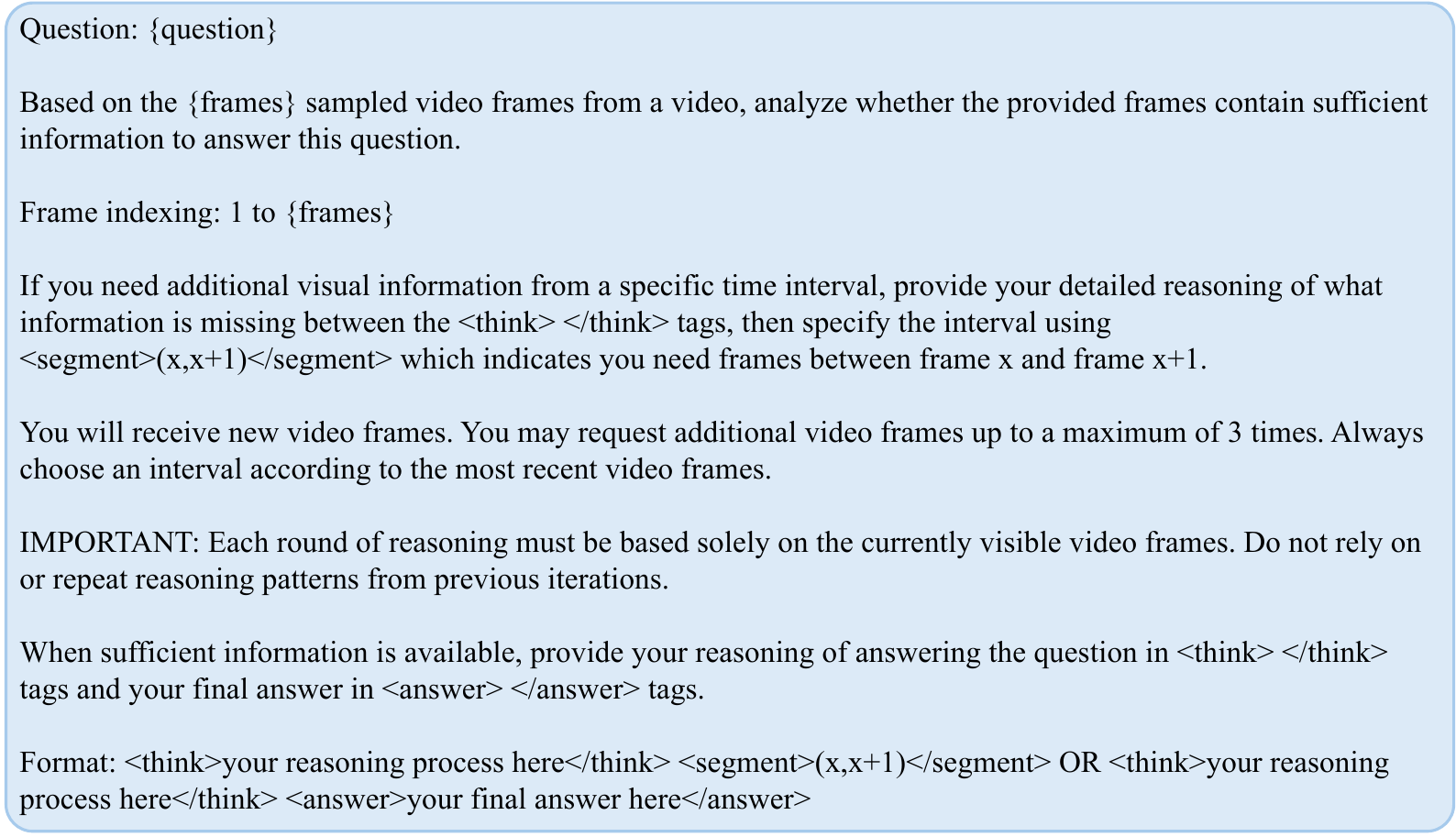}
    \caption{\textbf{Prompt template for the target model.} The target model is guided to perform temporal reasoning and either predict evidence segments for further exploration or generate final answers.}
    \label{fig:target_prompt}
\end{figure*}

\begin{figure*}[t]
    \centering
    \includegraphics[width=0.92\textwidth]{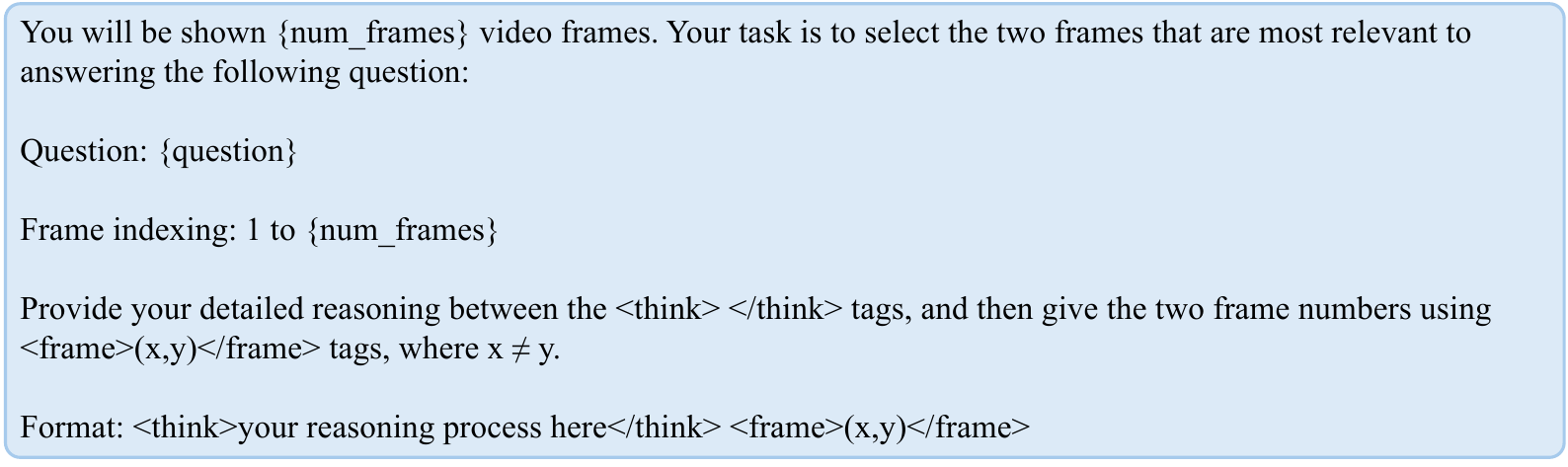}
    \caption{\textbf{Prompt template for the draft model.} The draft model is instructed to select the most informative and diverse frames from densely sampled temporal regions.}
    \label{fig:draft_prompt}
\end{figure*}

\noindent \textbf{Prompt templates for dual models.} The prompt templates for the target model (temporal segment prediction and answer generation) and the draft model (salient frame selection from dense samples) are illustrated in Figure~\ref{fig:target_prompt} and Figure~\ref{fig:draft_prompt}, respectively.

\noindent \textbf{SpecTemp-80K dataset statistics.} Figure~\ref{fig:figDataStatistics_arxiv}(a) presents the video duration distribution, showing three distinct categories: short-form videos ($<$1 min, 32.4\%), medium-length videos (1-10 min, 51.8\%), and long-form videos ($>$10 min, 15.8\%). Figure~\ref{fig:figDataStatistics_arxiv}(b) illustrates the data source composition of SpecTemp-80K. The dataset comprises 80,142 video-question-answer triplets collected from 9 diverse sources, with LLaVA-Video (30.2\%), ActivityNet (17.6\%), and Ego4D (14.8\%) constituting the primary sources. This diverse composition ensures comprehensive coverage across varying temporal complexities and video domains, enabling robust training for dual-model collaborative reasoning.


\subsection{More Experimental Findings} 

\begin{table}[h]
\centering
\caption{Ablation study on maximum iteration number $T_{\max}$. We report accuracy (\%) on LongVideoBench and Video-Holmes, along with actual iterations executed and inference latency.}
\label{tab:iterations}
\resizebox{\linewidth}{!}{
\begin{tabular}{c|cc|cc}
\toprule
$\boldsymbol{T}_{\max}$ & \textbf{Video-Holmes \cite{cheng2025video}} & \textbf{LongVideoBench \cite{wu2024longvideobench}} & \textbf{Iterations} & \textbf{Latency (s)} \\
\midrule
1 & 42.8 & 54.7 & \textbf{0.9} & \textbf{1.5} \\
2 & 44.6 & 55.1 & 1.7 & 2.0 \\
3 (Ours) & \textbf{47.0} & 57.5 & 2.3 & 2.3 \\
4 & 46.7 & 57.3 & 3.1 & 2.9 \\
5 & 46.2 & \textbf{58.1} & 3.7 & 3.5 \\
\bottomrule
\end{tabular}
}
\end{table}

\noindent \textbf{Ablation on different iteration numbers.} We investigate the impact of maximum iteration number $T_{\max}$ on performance and efficiency. As shown in Table~\ref{tab:iterations}, accuracy improves steadily as $T_{\max}$ increases from 1 to 3, achieving 57.5\% on LongVideoBench (+2.8\%) and 47.0\% on Video-Holmes (+4.2\%). However, beyond $T_{\max}=3$, we observe diminishing returns and even performance degradation on Video-Holmes (46.2\% at $T_{\max}=5$), suggesting that excessive iterations may introduce redundant frames and impair reasoning coherence. The actual average iteration numbers (0.9 to 3.7) are consistently lower than $T_{\max}$, demonstrating effective early termination. Considering that $T_{\max}=3$ achieves the best Video-Holmes performance and competitive LongVideoBench accuracy with only 2.3s latency, we adopt this configuration as the optimal balance between performance and efficiency.


\begin{table}[t]
\centering
\caption{Ablation study on reward components. We evaluate the contribution of IoU reward $R_{\text{IoU}}$ for temporal localization in the target model and visual information gain reward $R_{\text{visual}}$ for frame selection in the draft model.}
\label{tab:rewards}
\resizebox{\linewidth}{!}{
\begin{tabular}{cc|cc}
\toprule
$\boldsymbol{R_{\text{IoU}}}$ \textbf{(Target)} & $\boldsymbol{R_{\text{visual}}}$ \textbf{(Draft)} & \textbf{Video-Holmes} \cite{cheng2025video} & \textbf{LongVideoBench \cite{wu2024longvideobench}} \\
\midrule
- & - & 42.1 & 53.1 \\
\checkmark & - & 45.5 & 55.8 \\
- & \checkmark & 44.7 & 54.6 \\
\checkmark & \checkmark & \textbf{47.0} & \textbf{57.5} \\
\bottomrule
\end{tabular}
}
\end{table}

\noindent \textbf{Ablation on reward components.} We conduct ablation studies to analyze the contribution of individual reward components in our reinforcement learning framework. As shown in Table~\ref{tab:rewards}, both the IoU reward $R_{\text{IoU}}$ for the target model and the visual information gain reward $R_{\text{visual}}$ for the draft model provide substantial improvements over the SFT baseline (53.1\% on LongVideoBench, 42.1\% on Video-Holmes). Individually, $R_{\text{IoU}}$ contributes larger gains (+2.7\% on LongVideoBench, +3.4\% on Video-Holmes) compared to $R_{\text{visual}}$ (+1.5\% on LongVideoBench, +2.6\% on Video-Holmes), indicating that accurate temporal localization is the primary bottleneck for long video understanding. Importantly, combining both rewards yields the best performance (57.5\% on LongVideoBench, 47.0\% on Video-Holmes), with improvements of +4.4\% and +4.9\% over baseline respectively, demonstrating that the two reward signals provide complementary supervision for effective dual-model collaboration.


\begin{table}[t]
\centering
\caption{Ablation study on frame allocation strategies (Initial+Per-Iter×Max-Iter) under fixed budgets of 16 and 64 frames. We report accuracy (\%) and inference latency.}
\label{tab:frame_allocation}
\resizebox{\linewidth}{!}{
\begin{tabular}{cccc}
\toprule
\textbf{Strategy} & \textbf{Video-Holmes \cite{cheng2025video}} & \textbf{LongVideoBench~\cite{wu2024longvideobench}} & \textbf{Latency (s)} \\
\hline
\multicolumn{4}{c}{\emph{\textbf{16-frames budget}}} \\
\hline
4+4×3 & 44.3 & 55.8 & \textbf{1.5} \\
10+2×3 (Ours) & \textbf{47.0} & \textbf{57.5} & 1.8 \\
13+1×3 & 45.6 & 56.2 & 2.0 \\
\hline
\multicolumn{4}{c}{\emph{\textbf{64-frames budget}}} \\
\hline
32+8×4 (Ours) & \textbf{47.8} & \textbf{61.4} & \textbf{4.7} \\
48+4×4 & 46.1 & 59.7 & 5.2 \\
56+2×4 & 46.9 & 60.2 & 5.5 \\
\bottomrule
\end{tabular}
}
\end{table}

\noindent \textbf{Frame allocation strategies under fixed budget.} To investigate the optimal frame allocation strategy under a fixed computational budget, we conduct ablation studies with different distributions of initial frames and per-iteration frames. As shown in Table~\ref{tab:frame_allocation}, we evaluate three configurations for both 16-frame and 64-frame budgets, where the strategy is denoted as Initial+Per-Iter×Max-Iter.

The results demonstrate that our default configurations consistently achieve the best performance across both settings. For the 16-frame budget, the 10+2×3 strategy attains 57.5\% on LongVideoBench and 47.0\% on Video-Holmes, outperforming both alternatives with only 1.8s latency. For the 64-frame budget, our 32+8×4 configuration achieves 61.4\% and 47.8\% respectively while maintaining the lowest latency (4.7s) among all 64-frame strategies.

These findings reveal an important trade-off in frame allocation strategies. Aggressive early sampling (4+4×3, 48+4×4) prioritizes many frames per iteration but suffers from insufficient global temporal context during initial reasoning, leading to suboptimal segment prediction. Conversely, conservative late sampling (13+1×3, 56+2×4) allocates excessive frames to initial observation but lacks the granularity needed for fine-grained verification in subsequent iterations. Our balanced approach allocates moderate initial frames for coarse temporal understanding while reserving sufficient budget for iterative refinement, effectively bridging global perception and local examination. This validates that our frame allocation strategy achieves an optimal equilibrium between temporal reasoning breadth and verification depth.

\begin{figure*}[!ht]
    \centering
        \includegraphics[width=0.9\textwidth]{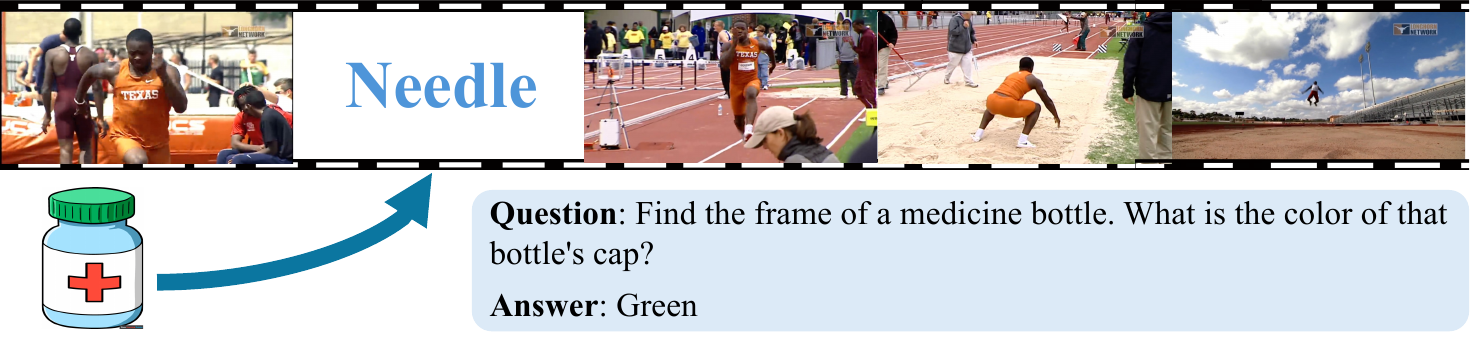}
     \caption{Illustration of the Visual Needle-In-A-Haystack evaluation. A single-frame visual question is inserted at varying positions within long video sequences to assess long-range retrieval capability.}
    \label{fig:visNIAH}
\end{figure*}
\begin{figure*}[!ht]
    \centering
        \includegraphics[width=0.95\textwidth]{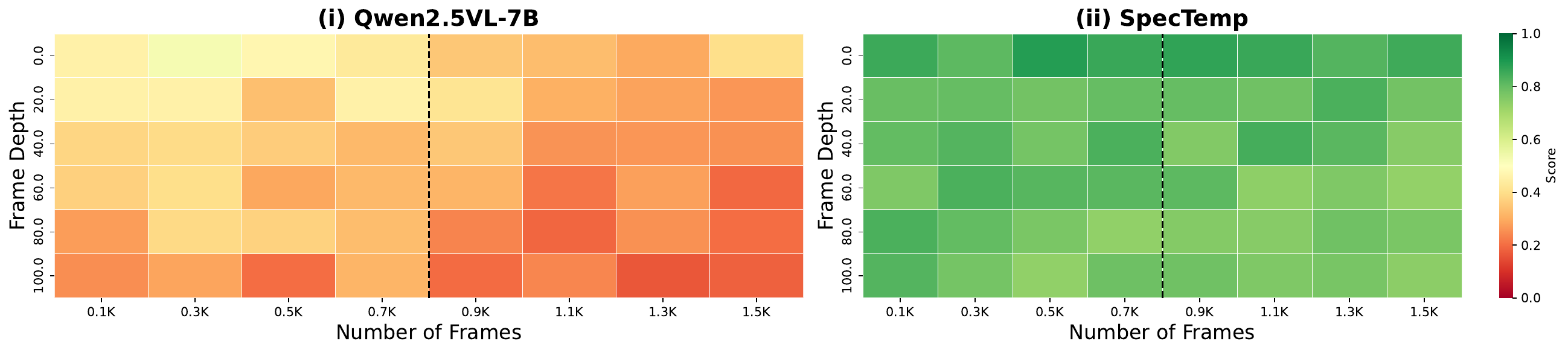}
     \caption{V-NIAH performance across frame counts and needle depths.}
    \label{fig:NIAH}
\end{figure*}

\noindent \textbf{Needle-in-a-Haystack evaluation.} To evaluate SpecTemp's ability to retrieve visual information from extremely long videos, we conduct a Visual Needle-In-A-Haystack (NIAH) test inspired by \cite{zhang2024long}. We insert single-frame visual questions at various positions into hour-long video haystacks sampled at 1 FPS. The needle images are designed to be counterfactual or counter-commonsense to ensure the model cannot rely on language priors alone.

As shown in Figure~\ref{fig:NIAH}, SpecTemp successfully retrieves needle information within videos containing up to 2000 frames, maintaining accuracy above 80\% across different needle positions. The dual-model collaboration (Target 7B + Draft 3B) achieves superior retrieval accuracy compared to using the 7B model alone, demonstrating that our speculative temporal reasoning framework maintains precise long-range visual information retrieval while significantly improving computational efficiency.

\subsection{Visualizations}

\begin{figure*}[t]
	\centering
        \includegraphics[width=0.9\textwidth]{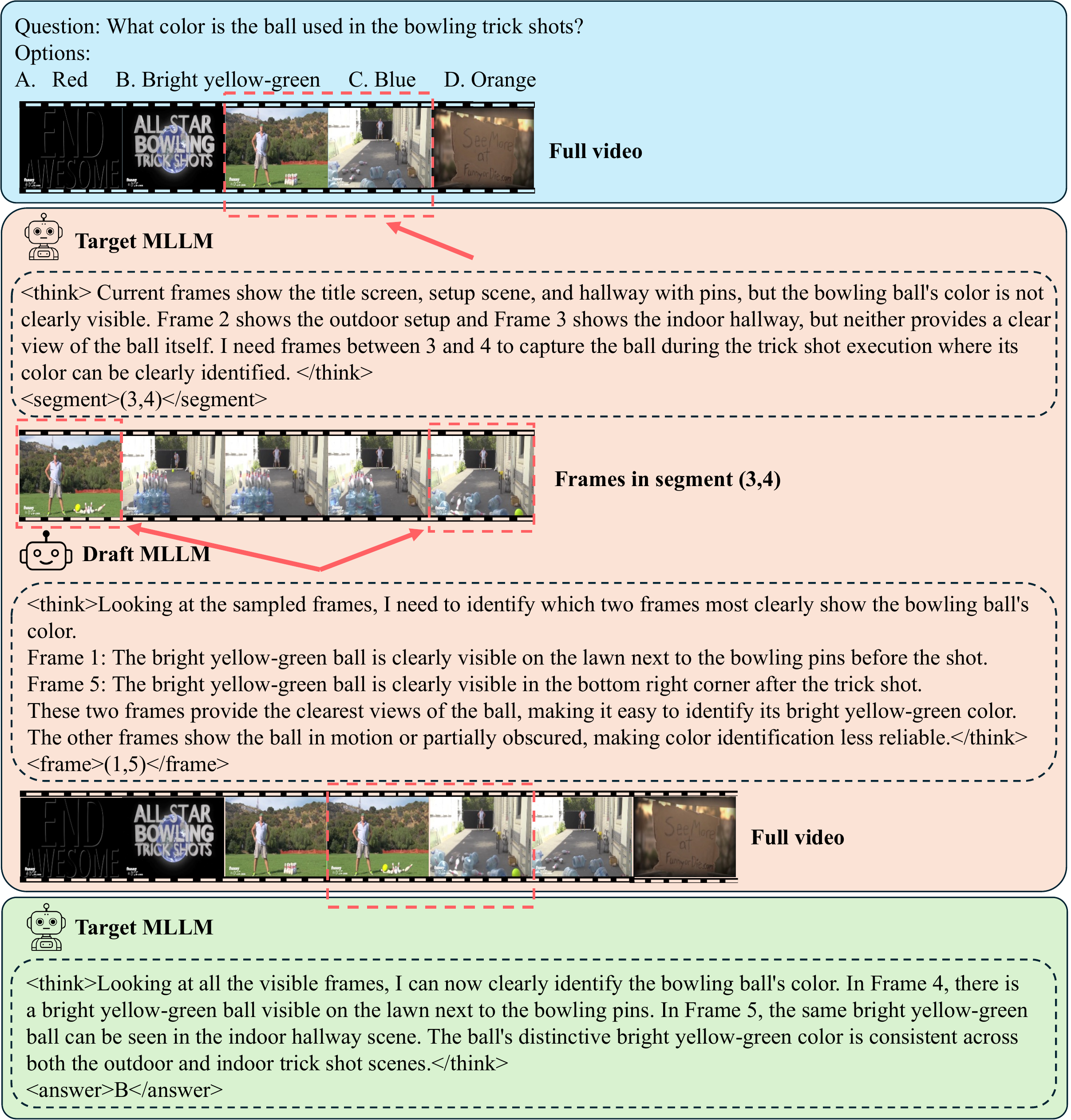}
	\caption{Qualitative case: 1 iteration for bowling ball color identification.
}
	\label{fig:vis_case_suppl1}
\end{figure*}

\begin{figure*}[t]
	\centering
        \includegraphics[width=0.9\textwidth]{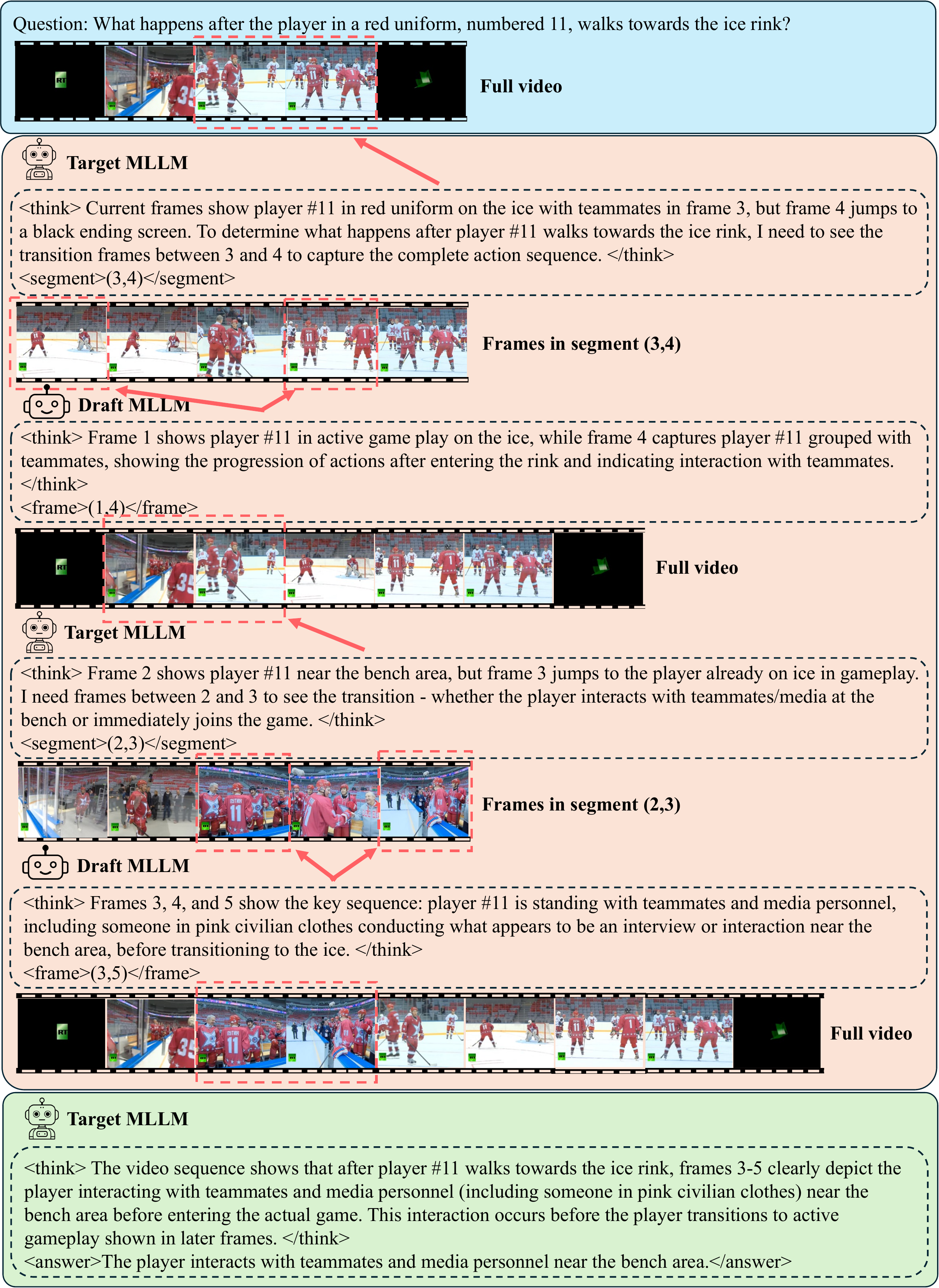}
	\caption{Qualitative case: 2 iterations for sports action sequence tracking.
}
	\label{fig:vis_case_suppl2}
\end{figure*}

\begin{figure*}[t]
	\centering
        \includegraphics[width=0.9\textwidth]{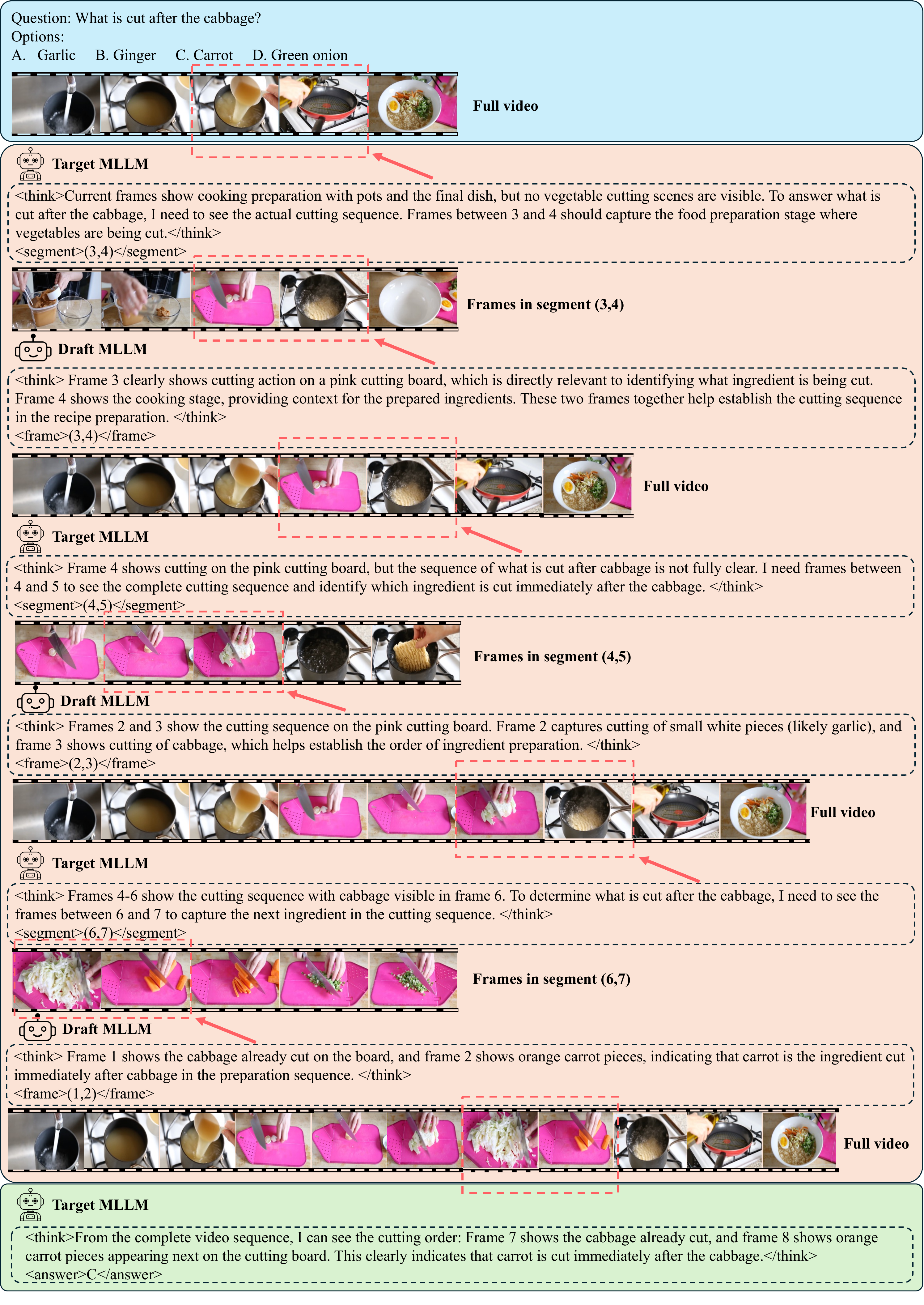}
	\caption{Qualitative case: 3 iterations for temporal ordering reasoning.}
	\label{fig:vis_case_suppl3}
\end{figure*}

We provide qualitative examples to illustrate SpecTemp's iterative speculation-verification process across different video understanding scenarios. Figures~\ref{fig:vis_case_suppl1} – Figures~\ref{fig:vis_case_suppl3} demonstrate how the target model progressively refines temporal segments while the draft model proposes salient frames, showcasing effective dual-model collaboration for long video reasoning.

\subsection{Limitations}

Despite the effectiveness of SpecTemp, we acknowledge several limitations that present opportunities for future research:

\noindent \textbf{Computational overhead during training.} Our dual-model framework requires training two separate MLLMs and performing joint reinforcement learning optimization, which increases computational costs compared to training a single model. The RL stage is particularly resource-intensive, requiring generation of multiple trajectory samples per training example for group reward computation. Future work could explore more efficient training strategies, such as parameter-efficient fine-tuning methods.

\noindent \textbf{Generalization to extremely long videos.} While SpecTemp handles hour-long videos effectively, its performance on videos exceeding 2-3 hours has not been extensively evaluated. For such extremely long videos, the number of potential temporal segments grows dramatically, and maintaining coherent reasoning across numerous iterations becomes challenging. Hierarchical reasoning strategies or memory mechanisms may be necessary for handling ultra-long videos.

{
    \small
    \bibliographystyle{ieeenat_fullname}
    \bibliography{references}
}


\end{document}